\documentclass[conference]{IEEEtran}
\IEEEoverridecommandlockouts
\usepackage{booktabs}
\usepackage{cite}
\usepackage{amsmath,amssymb,amsfonts}
\usepackage{algorithmic}
\usepackage{graphicx}
\usepackage{textcomp}
\usepackage{url}
\usepackage{float}
\usepackage{xcolor}
\usepackage{algorithm}
\usepackage{pifont} 
\def\BibTeX{{\rm B\kern-.05em{\sc i\kern-.025em b}\kern-.08em
    T\kern-.1667em\lower.7ex\hbox{E}\kern-.125emX}}
\begin{document}

\title{MpoxMamba: A Grouped Mamba-based Lightweight Hybrid Network for Mpox Detection\\
\thanks{This work is supported by NSF of Guangdong Province (No.2022A1515011044 and No.2023A1515010885), and the project of promoting research capabilities for key constructed disciplines in Guangdong Province (No.2021ZDJS028). '*' is corresponding author.}
}

\author{\IEEEauthorblockN{1\textsuperscript{st} Yubiao Yue}
\IEEEauthorblockA{\textit{School of Biomedical Engineering} \\
\textit{Guangzhou Medical University}\\
Guangzhou, China \\
jiche2020@126.com}
\and
\IEEEauthorblockN{2\textsuperscript{nd} Jun Xue}
\IEEEauthorblockA{\textit{Echo Tech} \\
Hefei, China \\
junxue.tech@gmail.com}
\and
\IEEEauthorblockN{3\textsuperscript{rd} Haihuang Liang}
\IEEEauthorblockA{\textit{School of Mathematics and Systems Science} \\
\textit{Guangdong Polytechnic Normal University}\\
Guangzhou, China \\
lianghhgdin@126.com}
\and
\IEEEauthorblockN{4\textsuperscript{th} Zhenzhang Li\textsuperscript{*}}
\IEEEauthorblockA{\textit{School of Mathematics and Systems Science} \\
\textit{Guangdong Polytechnic Normal University}\\
Guangzhou, China \\
zhenzhangli@gpnu.edu.cn}
\and
\IEEEauthorblockN{5\textsuperscript{th} Yufeng Wang}
\IEEEauthorblockA{\textit{Faculty of Computer Science and Information Technology} \\
\textit{Universiti Malaya}\\
Kuala Lumpur, Malaysia \\
fernandowangyf@gmail.com}
}

\maketitle

\begin{abstract}
Due to the lack of effective mpox detection tools, the mpox virus continues to spread worldwide and has once again been declared a public health emergency of international concern by the World Health Organization. Lightweight deep learning model-based detection systems are crucial to alleviate mpox outbreaks since they are suitable for widespread deployment, especially in resource-limited scenarios. However, the key to its successful application depends on ensuring that the model can effectively model local features and long-range dependencies in mpox lesions while maintaining lightweight. Inspired by the success of Mamba in modeling long-range dependencies and its linear complexity, we proposed a lightweight hybrid architecture called MpoxMamba for efficient mpox detection. MpoxMamba utilizes depth-wise separable convolutions to extract local feature representations in mpox skin lesions and greatly enhances the model's ability to model the global contextual information by grouped Mamba modules. Notably, MpoxMamba's parameter size and FLOPs are 0.77M and 0.53G, respectively. Experimental results on two widely recognized benchmark datasets demonstrate that MpoxMamba outperforms state-of-the-art lightweight models and existing mpox detection methods. Importantly, we developed a web-based online application to provide free mpox detection (\url{http://5227i971s5.goho.co:30290/}). The source codes of MpoxMamba are available at \url{https://github.com/YubiaoYue/MpoxMamba}.
\end{abstract}


\section{Introduction}
\begin{figure}
    \centering
    \includegraphics[width=1\linewidth]{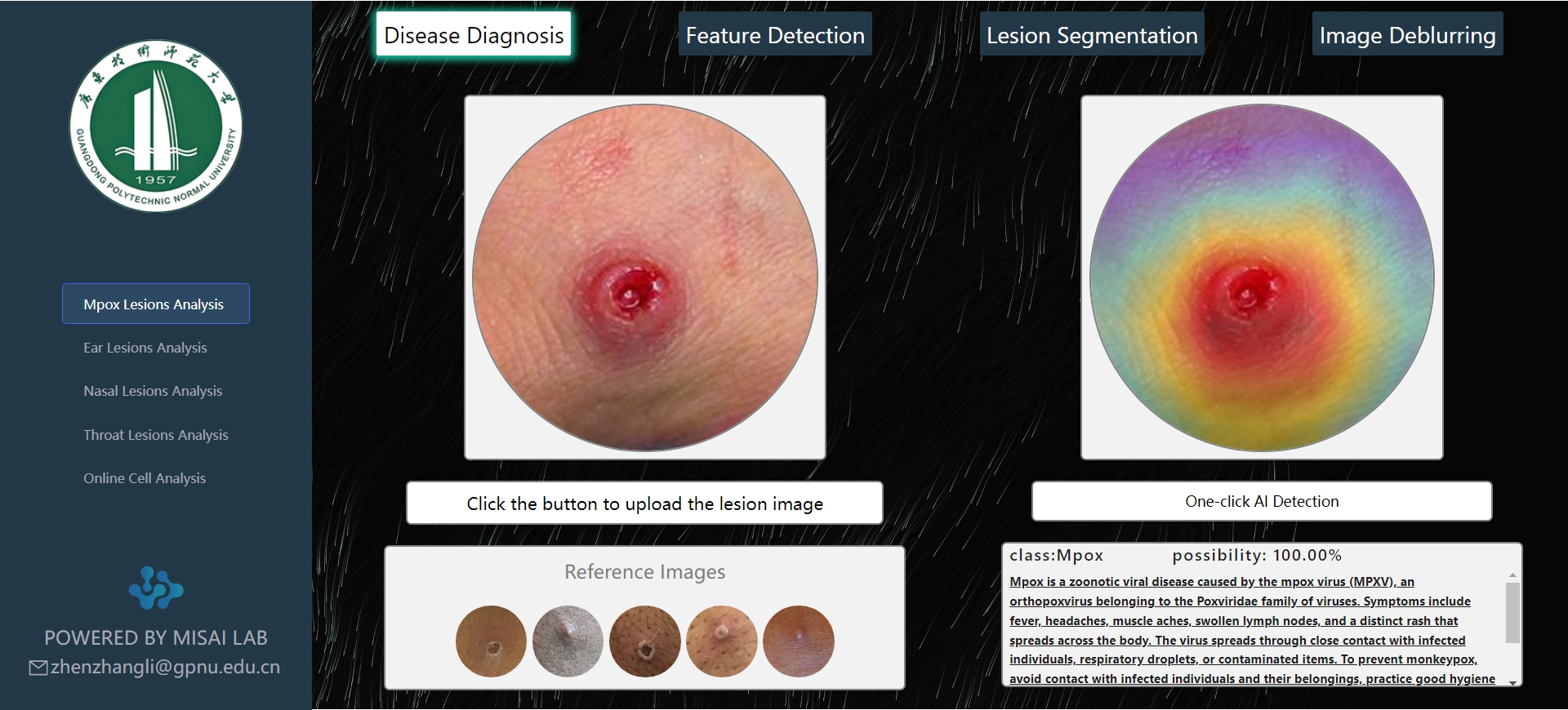}
    \caption{The overall page of online mpox detection application.}
    \label{fig1}
\end{figure}

\begin{figure*}
    \centering
    \includegraphics[width=1\textwidth]{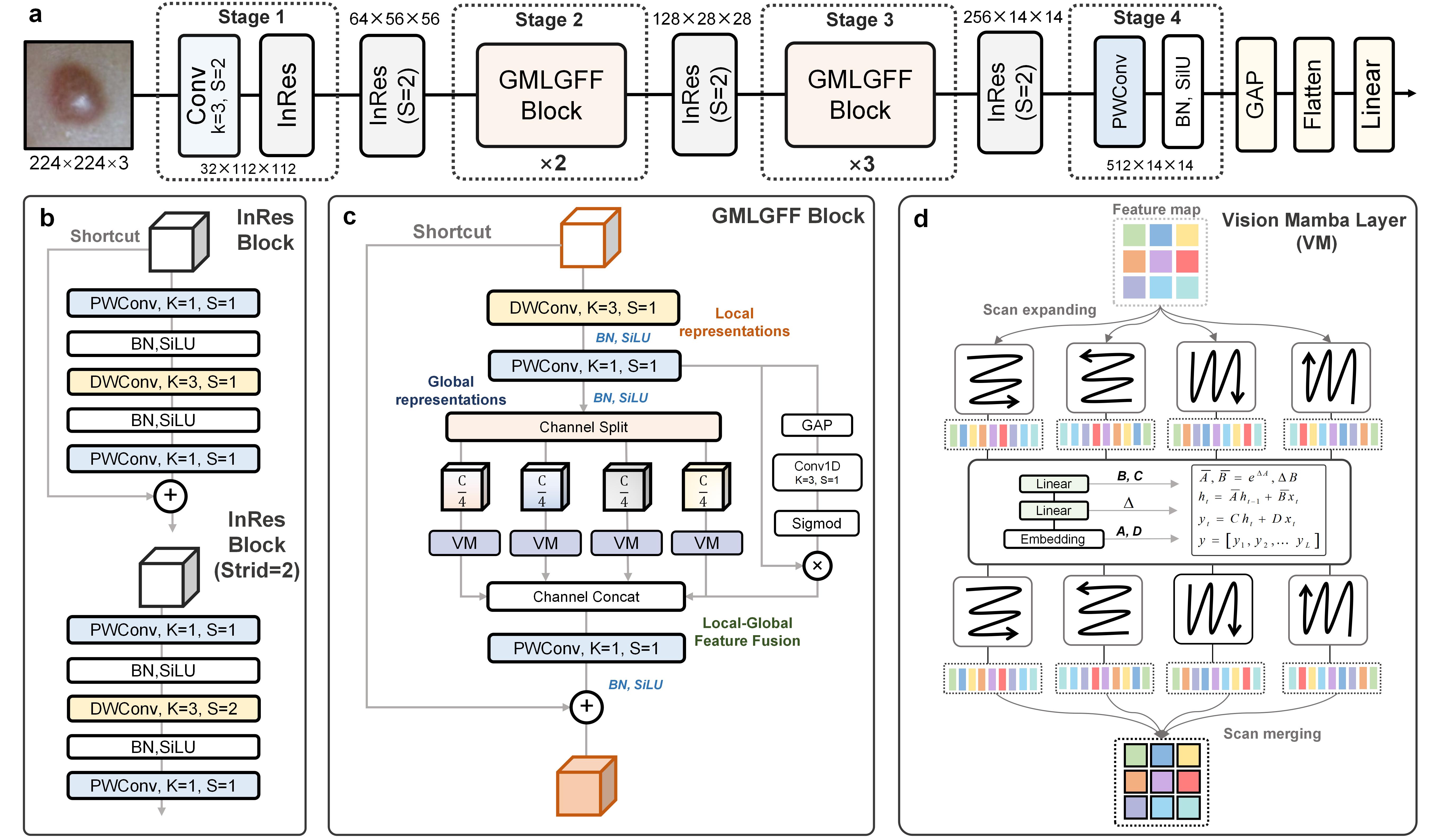}
    \caption{Detailed architecture of MpoxMamba and its internal components. GAP, BN, PWConv, DWConv represents global average pooling, batch normalization, point-wise convolution and depth-wise convolution, respectively. a: The overall architecture of MpoxMamba. b: The detailed architecture of InResBlock. c: The detailed architecture of GMLGFF Block. d: The detailed architecture of Vision Mamba Layer.}
    \label{fig2}
\end{figure*}
The fast-spreading mpox (monkeypox) poses a huge threat to global health \cite{ref1,ref2}. Over 120 countries have reported mpox between Jan 2022 and Aug 2024, with over 100,000 laboratory-confirmed cases reported and over 220 deaths among confirmed cases. Given the upsurge of mpox in the Democratic Republic of the Congo (DRC) and a growing number of countries in Africa, WHO declares mpox outbreak a public health emergency of international concern on August 14, 2024 \cite{ref3}, highlighting the urgent need for rapid and accurate mpox detection tools.

Although accurate, traditional mpox detection methods, such as Polymerase Chain Reaction testing, are often limited by equipment, cost, and time efficiency, which is not conducive to early detection and large-scale screening, especially in areas with limited medical resources \cite{ref4}. Deep learning systems based on skin lesion images have been proven to be a reliable solution for rapid and accurate mpox detection. Although most existing methods have successfully utilized convolutional neural networks (CNNs) to detect mpox \cite{ref5,ref6}, the inherent locality of the convolution operator limits the model's ability to capture long-range dependencies in skin lesion images. Long-range dependencies are a prerequisite for the formation of contextual information, which is crucial for accurate skin disease classification \cite{ref7}. Meanwhile, due to the intra-class variability and inter-class similarity in skin disease images, an effective mpox detection model should effectively integrate local features and global contextual information\cite{ref8,ref9}. To address this problem, some studies have used the Vision Transformer (ViT) to enhance the CNN-based model's ability to model long-distance dependencies \cite{ref10,ref11}. Unfortunately, ViT's detection performance relies on a large amount of pre-training data and the quadratic complexity of the self-attention mechanism results in expensive computational costs, which hinders the real-time application of mpox detection in large-scale screening scenarios. Most importantly, existing high-accuracy mpox detection models without exception have large parameter sizes and are not suitable for widespread deployment, especially on devices with limited computing and memory resources (Such as smartphones and development boards).

Recent studies have shown that Mamba (selective state space model) with linear computational complexity exhibits excellent performance in modeling long-distance  dependencies\cite{ref12}. Inspired by this, we hypothesize that Mamba can enhance lightweight CNN's ability to extract global contextual information, thereby promoting the widespread application of accurate and efficient mpox detection tools, which has important realistic meaning for epidemic prevention and control and global public health security. In this work, we investigated the potential of lightweight hybrid networks based on Mamba-CNN for mpox detection, aiming to achieve a good trade-off between detection performance, parameter size, and model complexity. To the best of our knowledge, this study represents the first attempt to use Mamba for intelligent mpox detection.

The contributions of this work can be summarized as follows: (1) We first proposed a lightweight hybrid network based on Mamba and CNN for efficient mpox detection, named MpoxMamba. The parameter size and model complexity (FLOPs) of MpoxMamba are only 0.77M and 0.53G. (2) We introduced a novel Grouped Mamba-based Local-Global Feature Fusion Block (GMLGFF). GMLGFF uses the grouped Mamba to enhance the model's ability to capture the long-range dependencies, helping the model perceive and aggregate scattered local features and global contextual information in mpox lesions effectively. By splitting feature maps and parallel processing, GMLGFF greatly reduces parameter size and model complexity. (3) Experimental results on two widely recognized benchmark datasets show that the MpoxMamba surpasses the state-of-the-art (SOTA) lightweight networks and existing methods. (4) A web-based online application was developed to provide free mpox detection services to the public in the epidemic area Fig. \ref{fig1}.

\section{The Proposed Method}
\subsection{Preliminaries}
Mamba relys on a classical continuous system that maps a one-dimensional input function or sequence, denoted as $x(t) \in \mathcal{R}$, through intermediate implicit states $h(t) \in \mathcal{R}^N$ to an output $y(t) \in \mathcal{R}$. The aforementioned process can be represented as a linear Ordinary Differential Equation (ODE)\cite{ref13,ref14,ref15}: 

\begin{equation}
\begin{aligned}
    &h^{\prime}(t) = \mathbf{A}h(t) + \mathbf{B}x(t) \\
    &y(t) = \mathbf{C}h(t)
\end{aligned}
\label{eq1}
\end{equation} where $\mathbf{A} \in \mathcal{R}^{N \times N}$ represents the state matrix, and $\mathbf{B} \in  \mathcal{R}^{N \times 1}$ and $\mathbf{C} \in  \mathcal{R}^{N \times 1}$ denote the projection parameters.

The Mamba discretize the continuous system to make it more suitable for deep learning. Concretely, Mamba introduce a timescale parameter $\mathbf{\Delta}$ to transform $\mathbf{A}$ and $\mathbf{B}$ into discrete parameters $\overline{\mathbf{A}}$ and $\overline{\mathbf{B}}$ using a fixed discretization rule. Typically, the zero-order hold (ZOH) is employed as the discretization rule and can be defined as follows:

\begin{equation}
\begin{aligned}
    &\overline{\mathbf{A}} = \textup{exp}(\mathbf{\Delta} \mathbf{A}) \\
    &\overline{\mathbf{B}} = (\mathbf{\Delta} \mathbf{A})^{-1}(\textup{exp}(\mathbf{\Delta} \mathbf{A}) - \mathbf{I})\cdot\mathbf{\Delta} \mathbf{B}
\end{aligned}
\label{eq2}
\end{equation}

After discretization, Equation \ref{eq1} utilizing a step size $\mathbf{\Delta}$ can be redefined as follows:

\begin{equation}
\begin{aligned}
    &h^{\prime}(t) = \overline{\mathbf{A}}h(t) + \overline{\mathbf{B}}x(t) \\
    &y(t) = \mathbf{C}h(t)
\end{aligned}
\label{eq:linear_recurrence}
\end{equation}
At the end of the process, the Mamba employ a global convolution to calculate the output: 
\begin{equation}
\begin{aligned}
    &\overline{K} = (\mathbf{C}\overline{\mathbf{B}}, \mathbf{C}\overline{\mathbf{AB}}, \ldots, \mathbf{C}\overline{\mathbf{A}}^{L-1}\overline{\mathbf{B}}) \\
    &y = x * \overline{\mathbf{K}}
\end{aligned}
\label{eq:global_convolution}
\end{equation}where $\overline{\mathbf{K}} \in \mathcal{R}^{L}$ represents a structured convolutional kernel, and $L$ denotes the length of the input sequence $x$.

\subsection{The Overall architecture of MpoxMamba}
Fig. \ref{fig2}a shows the overall framework of MpoxMamba. We first used a convolution layer and inverted residual (InRes) block \cite{ref16} (Fig. \ref{fig2}b) to capture low-level features such as the appearance, shape, and color of the rash image, resulting in a ($32 \times \frac{H}{2} \times \frac{W}{2}$) feature map. Then, an InRes block with stride=2 is used to quickly reduce the spatial dimension and optimize the extracted features. Stage 2 and Stage 3, constructed by stacked GMLGFF Blocks, are used to fully extract local features and global context information. The InRes Blocks (stride=2) are used to downsample the outputs of Stage2 and Stage3, outputting two feature maps with resolutions of ($128 \times \frac{H}{8} \times \frac{W}{8}$) and ($256 \times \frac{H}{16} \times \frac{W}{16}$), respectively. A point-wise convolution is used at the end of the network to map the feature map to a high-dimensional space ($512 \times \frac{H}{16} \times \frac{W}{16}$), enriching the feature representations. Finally, the adaptive average pooling and linear layers are used to detect mpox and non-mpox. The activation function and input size of MpoxMamba are set to SiLU and 224×224×3, respectively.

\begin{table*}[ht]
\centering
\caption{Performance comparison of MpoxMamba and reference models. Paras represents parameter size. Red and blue fonts indicate the best and second-best results, respectively. The “*” represents SOTA lightweight models.}
\begin{tabular}{ccccccccc}
\toprule
\textbf{Model}    & \textbf{Paras} & \textbf{FLOPs}    & \textbf{OA(MSLD)} & \textbf{Se(MSLD)} & \textbf{Sp(MSLD)} & \textbf{OA(MSID)} & \textbf{Se(MSID)} & \textbf{Sp(MSID)}       \\ \midrule
Mobileone-S0*\cite{ref20}       &4.27M	&1.07G	&73.58\%	&55.46\%	&88.12\%	&70.01\%	&57.45\%	&88.74\%  \\
MobileViT-XS*\cite{ref21}        &4.94M	&1.43G	&74.92\%	&70.36\%	&78.56\%	&72.45\%	&60.92\%	&89.94\%   \\
EfficientFormerV2-S1*\cite{ref22}   &5.74M	&0.66G	&71.92\%	&60.90\%	&80.90\%	&63.52\%	&51.85\%	&86.48\%   \\
FastViT-t8*\cite{ref23}    &\textcolor[RGB]{0,153,76} {3.26M}	&\textcolor{red} {0.53G}	&75.00\%	&66.64\%	&81.76\%	&73.36\%	&63.07\%	&90.19\%      \\
FasterNet-T1*\cite{ref24}   &6.32M	&0.85G	&77.19\%	&69.54\%	&83.36\%	&73.23\%	&62.34\%	&90.11\%  \\
MonkeyNet\cite{ref5}          &18.10M	&4.31G	& \textcolor[RGB]{0,153,76} {79.44\%}	&\textcolor{red} {77.36\%}	&81.04\%	&\textcolor[RGB]{0,153,76} {81.05\%}	&\textcolor[RGB]{0,153,76}{72.00}\%	&\textcolor[RGB]{0,153,76} {93.02\%}   \\
PoxNet22\cite{ref6}         &24.35M	&5.70G	&74.50\%	&67.64\%	&80.1\%	&72.09\%	&62.44\%	&89.89\%    \\
MpoxNet\cite{ref25}    &8.86M	&1.37G	&74.94\%	&67.46\%	&80.96\%	&76.51\%	&67.35\%	&91.28\%\\
ResNet50\cite{ref26}   &23.52M	&4.10G	&76.25\%	&74.18\%	&77.82\%	&74.82\%	&66.34\%	&90.85\% \\
SwinTransformer-T\cite{ref27}   &27.52M	&4.49G	&77.63\%	&70.64\%	&83.32\%	&75.32\%	&67.48\%	&91.33\% \\
ShuffleNetV2-X20 \cite{ref28}  &5.35M	& \textcolor[RGB]{0,153,76} {0.58G}	&72.33\%	&50.64\%	& \textcolor{red} {89.76\%}	&69.89\%	&59.70\%	&88.83\%    \\
EfficientNet-B1\cite{ref29}   &6.52M	&\textcolor[RGB]{0,153,76} {0.58G}	&71.02\%	&61.46\%	&78.64\%	&68.79\%	&56.36\%	&88.37\%  \\

 \midrule
\textcolor{red}{\textbf{MpoxMamba(Ours)}}    & \textcolor{red}{\textbf{0.77M}}    & \textcolor{red}{\textbf{0.53G}} & \textcolor{red}{\textbf{82.47\%}} & \textcolor[RGB]{0,153,76}{\textbf{75.46\%}} & \textcolor[RGB]{0,153,76}{\textbf{88.12\%}}            & \textcolor{red}{\textbf{81.71\%}}  & \textcolor{red}{\textbf{74.07\%}} & \textcolor{red}{\textbf{93.37\%}}\\ \bottomrule
\label{table1}
\end{tabular}
\end{table*}

\subsection{Vision Mamba Layer}
The Vision Mamba (VM) layer integrates a cross-scan module \cite{ref15} to assist the model in traversing the spatial domain of image feature maps from four corners all across the feature map to the opposite location, which ensures that each pixel in a feature map integrates information from all other locations in different directions, resulting in a global receptive field without increasing the linear computational complexity. The S6 Block\cite{ref14} is used to process sequences from all directions to ensure that the model can effectively model long-distance dependencies. Finally, the output features from the four directions are merged through a scan-merging operation to construct the final 2-D feature map, resulting in a final output of the same size as the input. Fig. \ref{fig2}d shows in detail the modeling mechanism of the VM layer.

\subsection{Grouped Mamba-based Local-Global Feature Fusion}
The proposed GMLGFF block (Fig. \ref{fig2}c) first employs depth-wise separable convolution (local representation module) to extract local feature representation without changing the size of the feature map. Then, the output of the local representation module ($C \times H \times W$) is evenly divided into four sub-feature maps ($\frac{C}{4} \times H \times W$), and four Vison Mamba layers are used to model the sub-feature maps to extract global feature representation (global representation module). In the feature fusion module, Efficient Channel Attention \cite{ref17} is used to adaptively adjust the output of the local representation module, thereby highlighting the key local features of mpox lesions while suppressing irrelevant information. Subsequently, the optimized local features are concatenated with the output of the global representation module along the channel dimension ($2C \times H \times W$), and the concatenate feature map is modeled using point-wise convolution, which ensures that the model can consider both local details and global context information and further improves the interaction between local-global feature representations. A shortcut connection is used to keep the original features, thereby reducing the risk of information loss in deeper layers.

\section{Experiments and Results}
\subsection{Dataset and Evaluation Metrics}
Two widely recognized mpox datasets are used in this work: Monkeypox Skin Lesion Dataset (MSLD) \cite{ref18,ref19} and Monkeypox Skin Images Dataset (MSID)\cite{ref5}. MSLD is a binary classification dataset containing 102 mpox images and 126 non-mpox images. MSID is a multi-class dataset containing 279 mpox images, 91 measles images, 107 chickenpox images, and 293 normal skin images. Here, based on the characteristics of medical images, we used overall accuracy (OA), sensitivity, and specificity as evaluation metrics.

\subsection{Implementation Details}
We did not apply any pre-training strategy or data augmentation during the experiment to demonstrate as much as possible that the model performance only benefits from the Mamba-CNN-based model. Five-fold cross-validation was used to fully evaluate the model's performance and generalization ability, reducing the risk of overfitting. During the training process, we set the epoch and batch size to 100 and 16 respectively, and used the AdamW optimizer (learning rate=0.0001, $\beta$1=0.9, $\beta$2=0.999 and weight decay=1e-4) and cross-entropy loss to optimize the model. To reflect the superiority of the MpoxMamba, some SOTA lightweight networks and existing methods were selected as reference models. To ensure fairness, the network architecture was the only different factor in the model comparison experiment.

\subsection{Comparison of Results}
Table \ref{table1} reports the results of various metrics of MpoxMamba and the reference models. Since MSLD is a binary classification dataset, we only calculate the sensitivity (abbreviated as Se) and specificity (abbreviated as Sp) of the model in detecting mpox. For MSID, we calculate the average sensitivity and average specificity of all categories. 

The results show that the OA of MpoxMamba reached 82.47\% (MSLD) and 81.71\% (MSID), respectively, reaching SOTA performance among all models. Compared with MonkeyNet, the OA of MpoxMamba increased by 2.17\% and 0.66\%, respectively. Except for Se (MSLD), MpoxMamba surpassed MonkeyNet in other metric results. It is worth noting that the parameter size and FLOPs of MpoxMamba are only 0.77M and 0.53G, respectively, which are 95.74\% and 87.70\% lower than MonkeyNet, respectively. Compared with the FasterNet-T1, MpoxMamba improves OA by 7.47\% and 8.35\% on MSLD and MSID, respectively, achieving a great performance improvement while maintaining lightweight. A comprehensive analysis of the results in Table 1 demonstrates that MpoxMamba has shown great competitiveness in various metrics and achieved the best trade-off between detection performance, parameter size and model complexity.

\subsection{Ablation Experiment}
We used MSLD to conduct ablation experiment. The entire ablation experiment started with the base model (the base model means that we removed the global representation module and the feature fusion module in GMLGFF Block). We analyzed the impact of the number of branches of Grouped Mamba on model performance, parameter size, and GFLOPs. Table \ref{table2} shows the ablation experiment results in detail. It can be seen from the results that the VM and Fusion modules significantly improve the OA of the basic model but inevitably increase the model complexity and parameter size. Fortunately, increasing the number of branches of Grouped Mamba improves the model accuracy, parameter size, and model FLOPs.

\begin{table}[ht]
\centering 
\caption{Performance changes of the MpoxMamba with different modules.} 
\resizebox{\linewidth}{!}{%
\begin{tabular}{@{}ccccccccc@{}}
\toprule
\textbf{Model} & \textbf{VM} & \textbf{Fusion} & \textbf{VM=2} & \textbf{VM=3}  & \textbf{VM=4} & \textbf{OA} & \textbf{GFLOPs} & \textbf{Paras} \\ \midrule
Basic Model    & \textcolor{red}{\ding{55}}    & \textcolor{red}{\ding{55}}    & \textcolor{red}{\ding{55}}   & \textcolor{red}{\ding{55}}    & \textcolor{red}{\ding{55}}     &72.18\%	&0.29G	&0.34M \\

Model2         & \textcolor[RGB]{0,153,76}{\checkmark}    & \textcolor{red}{\ding{55}}    & \textcolor{red}{\ding{55}}   & \textcolor{red}{\ding{55}}    & \textcolor{red}{\ding{55}}     &77.63\%	&0.69G	&0.96M \\

Model3         & \textcolor[RGB]{0,153,76}{\checkmark}    & \textcolor[RGB]{0,153,76}{\checkmark}    & \textcolor{red}{\ding{55}}   & \textcolor{red}{\ding{55}}    & \textcolor{red}{\ding{55}}     &80.39\%	&0.82G	&1.07M \\

Model4         & \textcolor[RGB]{0,153,76}{\checkmark}    & \textcolor[RGB]{0,153,76}{\checkmark}    & \textcolor[RGB]{0,153,76}{\checkmark}   & \textcolor{red}{\ding{55}}    & \textcolor{red}{\ding{55}}     &82.03\%	&0.63G	&0.87M \\

Model5         & \textcolor[RGB]{0,153,76}{\checkmark}    & \textcolor[RGB]{0,153,76}{\checkmark}    & \textcolor[RGB]{0,153,76}{\checkmark}   & \textcolor[RGB]{0,153,76}{\checkmark}    & \textcolor{red}{\ding{55}}    &81.72\%	&0.56G	&0.81M \\

MpoxMamba      & \textcolor[RGB]{0,153,76}{\checkmark}    & \textcolor[RGB]{0,153,76}{\checkmark}    & \textcolor[RGB]{0,153,76}{\checkmark}   & \textcolor[RGB]{0,153,76}{\checkmark}    & \textcolor[RGB]{0,153,76}{\checkmark}     &82.47\%	&0.53G	&0.77M \\ \bottomrule
\end{tabular}
}
\label{table2}
\end{table}

\subsection{A Web-based Mpox Detection Application}
To promote the practical application of deep learning technology, we utilized the MpoxMamba to develop a web-based online application (Fig. \ref{fig1}). Users only need to upload a pre-taken rash image, and they can obtain real-time diagnosis results through the Internet. In order to improve the security of the application, we also used Grad-CAM \cite{ref30} to visually explain the model's decision-making results. In addition, medical knowledge and medical advice on mpox and non-mpox skin diseases and reference images are displayed on the result page to raise public health awareness. We hope that this work can provide the best possible help to alleviate mpox outbreak.

\section{Conclusion}
In summary, this work proposed a lightweight Mamba-CNN hybrid model that can efficiently detect mpox lesions in rash images. Compared with existing methods, MpoxMamba achieves SOTA performance while maintaining the lowest parameter size and model complexity. Future research will focus on exploring effective data augmentation techniques and pre-training strategies (such as self-supervised learning and transfer learning) to further improve the performance of MpoxMamba.

\vspace{12pt}

\clearpage
\begin{thebibliography}{00}

\bibitem{ref1}
S.~K. Saxena, S.~Ansari, V.~K. Maurya, S.~Kumar, A.~Jain, J.~T. Paweska, A.~K. Tripathi, and A.~S. Abdel-Moneim, ``Re-emerging human monkeypox: a major public-health debacle,'' {\em Journal of medical virology}, vol.~95, no.~1, p.~e27902, 2023.

\bibitem{ref2}
K.~Mercy, B.~Tibebu, M.~Fallah, N.~R. Faria, N.~Ndembi, and Y.~K. Tebeje, ``Mpox continues to spread in africa and threatens global health security,'' {\em Nature Medicine}, vol.~30, no.~5, pp.~1225--1226, 2024.

\bibitem{ref3}
W.~H. Organization, ``Mpox,'' 2024.
\newblock [Online]. Available: \url{https://www.who.int/news-room/fact-sheets/detail/mpox}.

\bibitem{ref4}
S.~K. Patel, J.~Rana, A.~Agrawal, N.~K. Channabasappa, A.~K. Niranjan, and T.~B. Emran, ``Mpox and the need for improved diagnostics--correspondence,'' {\em Annals of Medicine and Surgery}, vol.~85, no.~4, pp.~1323--1324, 2023.

\bibitem{ref5}
D.~Bala, M.~S. Hossain, M.~A. Hossain, M.~I. Abdullah, M.~M. Rahman, B.~Manavalan, N.~Gu, M.~S. Islam, and Z.~Huang, ``Monkeynet: A robust deep convolutional neural network for monkeypox disease detection and classification,'' {\em Neural Networks}, vol.~161, pp.~757--775, 2023.

\bibitem{ref6}
F.~Yasmin, M.~M. Hassan, M.~Hasan, S.~Zaman, C.~Kaushal, W.~El-Shafai, and N.~F. Soliman, ``Poxnet22: A fine-tuned model for the classification of monkeypox disease using transfer learning,'' {\em Ieee Access}, vol.~11, pp.~24053--24076, 2023.

\bibitem{ref7}
X.~He, E.-L. Tan, H.~Bi, X.~Zhang, S.~Zhao, and B.~Lei, ``Fully transformer network for skin lesion analysis,'' {\em Medical Image Analysis}, vol.~77, p.~102357, 2022.

\bibitem{ref8}
L.~Tan, H.~Wu, J.~Xia, Y.~Liang, and J.~Zhu, ``Skin lesion recognition via global-local attention and dual-branch input network,'' {\em Engineering Applications of Artificial Intelligence}, vol.~127, p.~107385, 2024.

\bibitem{ref9}
X.~Huo, G.~Sun, S.~Tian, Y.~Wang, L.~Yu, J.~Long, W.~Zhang, and A.~Li, ``Hifuse: Hierarchical multi-scale feature fusion network for medical image classification,'' {\em Biomedical Signal Processing and Control}, vol.~87, p.~105534, 2024.

\bibitem{ref10}
S.~Maqsood, R.~Dama{\v{s}}evi{\v{c}}ius, S.~Shahid, and N.~D. Forkert, ``Mox-net: Multi-stage deep hybrid feature fusion and selection framework for monkeypox classification,'' {\em Expert Systems with Applications}, vol.~255, p.~124584, 2024.

\bibitem{ref11}
G.~Yolcu~Oztel, ``Vision transformer and cnn-based skin lesion analysis: classification of monkeypox,'' {\em Multimedia Tools and Applications}, vol.~83, no.~28, pp.~71909--71923, 2024.

\bibitem{ref12}
L.~Zhu, B.~Liao, Q.~Zhang, X.~Wang, W.~Liu, and X.~Wang, ``Vision mamba: Efficient visual representation learning with bidirectional state space model,'' in {\em Proceedings of the 41st International Conference on Machine Learning} (R.~Salakhutdinov, Z.~Kolter, K.~Heller, A.~Weller, N.~Oliver, J.~Scarlett, and F.~Berkenkamp, eds.), vol.~235 of {\em Proceedings of Machine Learning Research}, pp.~62429--62442, PMLR, 21--27 Jul 2024.

\bibitem{ref13}
J.~Ruan and S.~Xiang, ``Vm-unet: Vision mamba unet for medical image segmentation,'' {\em arXiv preprint arXiv:2402.02491}, 2024.

\bibitem{ref14}
A.~Gu and T.~Dao, ``Mamba: Linear-time sequence modeling with selective state spaces,'' {\em arXiv preprint arXiv:2312.00752}, 2023.

\bibitem{ref15}
Y.~Liu, Y.~Tian, Y.~Zhao, H.~Yu, L.~Xie, Y.~Wang, Q.~Ye, and Y.~Liu, ``Vmamba: Visual state space model,'' {\em arXiv preprint arXiv:2401.10166}, 2024.

\bibitem{ref16}
M.~Sandler, A.~Howard, M.~Zhu, A.~Zhmoginov, and L.-C. Chen, ``Mobilenetv2: Inverted residuals and linear bottlenecks,'' in {\em Proceedings of the IEEE conference on computer vision and pattern recognition}, pp.~4510--4520, 2018.

\bibitem{ref17}
Q.~Wang, B.~Wu, P.~Zhu, P.~Li, W.~Zuo, and Q.~Hu, ``Eca-net: Efficient channel attention for deep convolutional neural networks,'' in {\em Proceedings of the IEEE/CVF conference on computer vision and pattern recognition}, pp.~11534--11542, 2020.

\bibitem{ref18}
S.~N. Ali, M.~T. Ahmed, T.~Jahan, J.~Paul, S.~S. Sani, N.~Noor, A.~N. Asma, and T.~Hasan, ``A web-based mpox skin lesion detection system using state-of-the-art deep learning models considering racial diversity,'' {\em Biomedical Signal Processing and Control}, vol.~98, p.~106742, 2024.

\bibitem{ref19}
S.~N. Ali, M.~T. Ahmed, J.~Paul, T.~Jahan, S.~Sani, N.~Noor, and T.~Hasan, ``Monkeypox skin lesion detection using deep learning models: A feasibility study,'' {\em arXiv preprint arXiv:2207.03342}, 2022.

\bibitem{ref20}
P.~K.~A. Vasu, J.~Gabriel, J.~Zhu, O.~Tuzel, and A.~Ranjan, ``Mobileone: An improved one millisecond mobile backbone,'' in {\em Proceedings of the IEEE/CVF conference on computer vision and pattern recognition}, pp.~7907--7917, 2023.

\bibitem{ref21}
S.~Mehta and M.~Rastegari, ``Mobilevit: Light-weight, general-purpose, and mobile-friendly vision transformer,'' in {\em International Conference on Learning Representations}, 2022.

\bibitem{ref22}
Y.~Li, J.~Hu, Y.~Wen, G.~Evangelidis, K.~Salahi, Y.~Wang, S.~Tulyakov, and J.~Ren, ``Rethinking vision transformers for mobilenet size and speed,'' in {\em Proceedings of the IEEE/CVF International Conference on Computer Vision}, pp.~16889--16900, 2023.

\bibitem{ref23}
P.~K.~A. Vasu, J.~Gabriel, J.~Zhu, O.~Tuzel, and A.~Ranjan, ``Fastvit: A fast hybrid vision transformer using structural reparameterization,'' in {\em Proceedings of the IEEE/CVF International Conference on Computer Vision}, pp.~5785--5795, 2023.

\bibitem{ref24}
J.~Chen, S.-h. Kao, H.~He, W.~Zhuo, S.~Wen, C.-H. Lee, and S.-H.~G. Chan, ``Run, don't walk: chasing higher flops for faster neural networks,'' in {\em Proceedings of the IEEE/CVF conference on computer vision and pattern recognition}, pp.~12021--12031, 2023.

\bibitem{ref25}
J.~Sun, B.~Yuan, Z.~Sun, J.~Zhu, Y.~Deng, Y.~Gong, and Y.~Chen, ``Mpoxnet: dual-branch deep residual squeeze and excitation monkeypox classification network with attention mechanism,'' {\em Frontiers in Cellular and Infection Microbiology}, vol.~14, p.~1397316, 2024.

\bibitem{ref26}
K.~He, X.~Zhang, S.~Ren, and J.~Sun, ``Deep residual learning for image recognition,'' in {\em Proceedings of the IEEE conference on computer vision and pattern recognition}, pp.~770--778, 2016.

\bibitem{ref27}
Z.~Liu, Y.~Lin, Y.~Cao, H.~Hu, Y.~Wei, Z.~Zhang, S.~Lin, and B.~Guo, ``Swin transformer: Hierarchical vision transformer using shifted windows,'' in {\em Proceedings of the IEEE/CVF international conference on computer vision}, pp.~10012--10022, 2021.

\bibitem{ref28}
N.~Ma, X.~Zhang, H.-T. Zheng, and J.~Sun, ``Shufflenet v2: Practical guidelines for efficient cnn architecture design,'' in {\em Proceedings of the European conference on computer vision (ECCV)}, pp.~116--131, 2018.

\bibitem{ref29}
M.~Tan and Q.~Le, ``{E}fficient{N}et: Rethinking model scaling for convolutional neural networks,'' in {\em Proceedings of the 36th International Conference on Machine Learning} (K.~Chaudhuri and R.~Salakhutdinov, eds.), vol.~97 of {\em Proceedings of Machine Learning Research}, pp.~6105--6114, PMLR, 09--15 Jun 2019.

\bibitem{ref30}
R.~R. Selvaraju, M.~Cogswell, A.~Das, R.~Vedantam, D.~Parikh, and D.~Batra, ``Grad-cam: Visual explanations from deep networks via gradient-based localization,'' in {\em Proceedings of the IEEE international conference on computer vision}, pp.~618--626, 2017.





\end{thebibliography}
\end{document}